\documentclass[review]{elsarticle}

\usepackage{lineno,hyperref}
\usepackage[utf8]{inputenc} 
\usepackage{graphicx}
\usepackage{amssymb}
\usepackage{amsmath}
\usepackage{amsfonts}
\usepackage{graphicx}
\usepackage{svg}
\usepackage{subfig} 
\usepackage{xcolor}
\usepackage{booktabs}
\usepackage{soul}
\usepackage{multirow}
\usepackage{afterpage,lscape}

\journal{Expert Systems with Applications}

\biboptions{authoryear}

\begin{document}

\begin{frontmatter}

\title{A Novel Semi-supervised Framework for Call Center Agent Malpractice Detection via Neural Feature Learning}

\author[a]{Leonardo Obinna Iheme}
\author[a]{Şükrü Ozan\fnref{myfootnote}}
\address[a]{Folkart Towers B Block, $36^{th}$ Floor, Office 3609, Bayraklı, İzmir, Turkey\\phone: +90 232 441 40 11}
\address{\textit {\{leonardoiheme, sukruozan\}@adresgezgini.com}}
\fntext[myfootnote]{Corresponding author}

\begin{abstract}
This work presents a practical solution to the problem of call center agent malpractice. A semi-supervised framework comprising of non-linear power transformation, neural feature learning and k-means clustering is outlined. We put these building blocks together and tune the parameters so that the best performance was obtained. The data used in the experiments is obtained from our in-house call center. It is made up of recorded agent-customer conversations which have been annotated using a convolutional neural network based segmenter. The methods provided a means of tuning the parameters of the neural network to achieve a desirable result. We show that, using our proposed framework, it is possible to significantly reduce the malpractice classification error of a k-means-only clustering model which would serve the same purpose. Additionally, by presenting the amount of silence per call as a key performance indicator, we show that the proposed system has enhanced agents performance at our call center since deployment.
\end{abstract}

\end{frontmatter}

\section{Introduction}\label{sec:intro}

The importance of telephony as a communication channel in marketing has been on the rise since the late 80s \citep{Marshall:1988} to such an extent that today, telemarketing is an integral part of marketing strategy work flows of most business organizations. In addition to telemarketing, for customer relationship management, telephony has proven to be the most effective customer retention strategy, guaranteeing satisfaction and loyalty \citep{Feinberg:2000}. Even though other channels such as social media platforms have recently attracted the attention of researchers and practitioners \citep{Iankova:2018}, the call center has remained the most effective customer relationship management channel. Hence, the need to optimize the performance cannot be over-emphasized.

Customer dissatisfaction arising from call center interactions often stem from agents' malpractices which ultimately leads to churn and loss of revenue. From our experience and observations over the years, we have realized that a major constituent of call center agent malpractice is the amount or pattern of silence in a call. It may be for an extended duration or in the form of specific unique patterns like in fax tones or interactive voice response systems \citep{Ozan:2019}. As a result, detecting, quantifying and analyzing silence patterns is key in detecting malpractices and providing valuable feedback to call center agents.

The fairly recent upsurge in compute capabilities and availability of data has brought about rapid improvements in supervised learning tasks in the form of deep learning. Of all the challenges of deep learning, the availability of labelled data is arguably at the forefront. The acquisition of labelled data often requires a skilled human however, this can be expensive and given the huge amount of labelled data required to properly train a deep neural network, labelling might be infeasible. Besides the cost associated with labelling, there exists the issue of reliability \citep{Armstrong:1997} wherein expert labels are subjective to an unknown number of variables.

Through the use of unsupervised learning algorithms, one may discover patterns in unlabeled data. Unsupervised learning algorithms have been instrumental in data exploration \citep{girolami:1998,steiger:2016}, segmentation \citep{Ng:2016,Chung:2016} and a host of other tasks. While the performance of the various algorithms have been impressive, the design of appropriate performance metrics has remained a somewhat challenging task. In fact, most available unsupervised learning performance metrics still require knowledge of the ground truth classes \citep{Hubert:1985,Vinh:2010,Rosenberg:2007}. It is however possible to combine the best of both worlds (supervised learning and unsupervised learning) in the form of semi-supervised learning.

Semi-supervised learning involves training and validating an unsupervised model such as k-means, with a large amount of unlabeled data and supervised learning performance metrics respectively. It is applicable in scenarios where manually labeling data is not feasible. In our use case for instance, the data used in the experiments consists of over 180,000 call center records translating to a total duration of over 540 hours. Labelling the data would involve listening to every call and classifying it accordingly. Not only will this process be time consuming but it will also be quite expensive. The unsupervised component of semi-supervised learning requires some assumptions or domain knowledge of the data. Assumptions such as the number of classes present in the data or the underlying distribution of the data need to be enforced.

To improve the performance of machine learning algorithms in general, various forms of feature engineering are carried out: feature selection \citep{Yan:2015}, scaling and normalization \citep{Stolcke:2008}, transformation \citep{Yeo:2000} and more recently, feature learning with neural networks \citep{Yu:2013}. Typically, in feature learning with neural networks, a non-classifying neural network is cascaded with a classifier then the parameters of the network are adjusted by the loss of the classifier. The role of the neural network is to extract more discriminative features from the raw features. As shown in \citep{Yu:2013}, through speech recognition, a better performance can be achieved with the extracted features. Although such a set-up increases the complexity of the system, it has the advantage of providing tunable hyper-parameters that affect the performance of the classifier. In a semi-supervised scenario, where the classifier has no hyper-parameters to be tuned, the neural network can intrinsically provide that capability while learning the best features possible.

In essence, using semi-supervised learning, we aim to detect call center agent malpractices and by applying feature learning, we aim to improve the performance of the system. Our proposed solution does not take the human (quality control personnel) out of the loop. We intend for the system to send notifications of possible agent malpractices to the quality control manager. As a result, for the machine learning algorithms, non-malpractices are considered to be the negative class while malpractices are the positive class. Our system ensures that true negatives and false positives are checked by the manager. Thus, our priority is to minimize the false negatives produced by the system. Accordingly, in this use case, the performance of the model is assessed by its ability to maximize the recall score \citep{Goutte:2005}.

\section{Agent Malpractice}
Malpractice, within the scope of agent-customer interaction over the phone involves practices which do not conform to the organization's stipulated rules, these are  practices which lead to revenue slippage and customer churn. These include but are not limited to extensive silence periods in a call, deliberately placing calls to fax lines as described in \citep{Ozan:2019} or interactive voice response (IVR) systems and keeping customers on hold unnecessarily for extended periods.

It is a known fact that humans can easily classify unstructured data (audio in this case). It however becomes an expensive venture when the amount of data is large and the time to classify is limited. To give some perspective, at our call center, 20 agents with one supervisor, generate $\sim$1000 calls per day ($\sim$ 50 calls per agent). Assuming the average duration of a call is 120 seconds, it will take over 33 hours to listen to the calls of a single day. Therefore, it will be impossible for the supervisor to listen to all the calls generated per day. Accordingly, the aim of this study is to automatically detect malpractices and notify the supervisor. It is the responsibility of the supervisor to make the final decision on the detected malpractices and act accordingly. 

\section{Data}\label{sec:data}
The data used in this work is a sample of 180,000 call center conversations drawn from a population of over a million call records. Calls are recorded and saved in a compressed \textit{.gsm} format with a sampling rate of 8000 Hertz. The sample calls were selected from a period of time when agent malpractice was at its peak at the organization. It was found that  at least 3\% of the calls in that period constituted some sort of malpractice on the agents' part. The total duration of the sample is {$\sim$} 540 hours.

Of the 180,000 calls, 3,885 were listened to and labeled by experts. 716 calls were labeled malpractice while 3169 were non-malpractice. This makes up our validation set, while the rest of the data constitutes the training set.

\section{Building Blocks}\label{sec:method}
In this section, we outline and explain the individual components which constitute the proposed framework.

\subsection{CNN-based Feature Extraction}
\begin{figure}[]
 \centering
 \includegraphics[width=\textwidth]{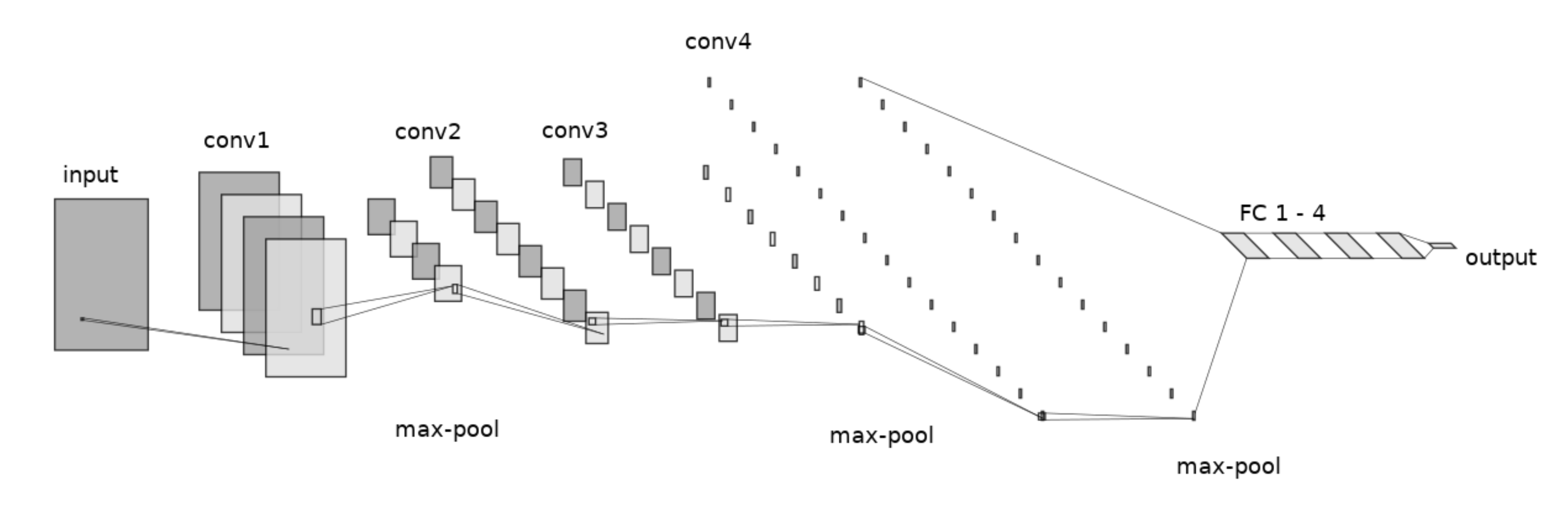}
 \caption{Representative figure of convolutional network used for segmenting speech, music and noise. Note the diagram is not drawn to scale.}
 \label{fig:cnn}
\end{figure}

From each recorded call, four features are extracted; namely percentage of speech, percentage of music, percentage of silence and percentage of noise. To obtain these, a simple automatic energy thresholding algorithm is first used to segment frames associated with low energy \citep{Doukhan:2018}, considered as silence. This is followed by MFCC feature extraction as specified in \citep{Doukhan:2017}. The MFCC features are fed as input to a CNN architectureto to segment speech, music and noise. The network, which is a modification of the work presented in \citep{Doukhan:2017}, consists of eight layers: four convolutional layers and four fully connected layers each followed by max-pooling and drop-out respectively. Rectified linear unit (ReLU) activation functions are used between layers and at the output, there are three neurons normalized by a softmax function. A summary of the CNN architecture is represented in Figure \ref{fig:cnn}. The model has a total of 788,803 parameters of which 785,603 of them are trainable. 

The final output of the segmentation tool is the starting time points, ending time points and the labels of the corresponding homogeneous segments respectively. From the start and stop times, the percentage of each unique segment is calculated accordingly. Finally, four features are obtained for each record.

\subsection{Feature Transformation}
\label{subsec:feat_trans}
Feature transformation is a  pre-processing step which aims to improve the performance of machine learning algorithms. In this study, we experiment with two feature transformations: the Z-score normalization and power transform. The goal of both transformations is to, as much as possible "Gaussianize" the data since this is a desirable property.

The Z-score normalization,  which is a linear transform, as shown in Equation \ref{eq:standard}, transforms each feature to a standard Gaussian distribution by removing the mean and scaling the data to unit variance. This puts each feature on the same scale and removes possible magnitude biases that might exist. The major drawback of this transformation is the fundamental assumption that the distribution is Gaussian.

\begin{equation}
    z_{i}= \frac{x_{i} -  \overline{x} }{s}
\label{eq:standard}
\end{equation}
where $\overline{x}$ is the sample mean and $s$ is the sample standard deviation.

Power transforms are a family of parametric, monotonic nonlinear functions that aim to map data from any distribution to as close to a Gaussian distribution as possible in order to stabilize variance and minimize skewness \citep{scikit-learn}. Specifically, in this work, we employ the Yeo–Johnson \citep{Yeo:2000} transformation which is an extension of the Box–Cox transformation \citep{Box:1964}. The Box–Cox transform family of functions is given in Equation \ref{eq:boxcox}.
\begin{equation}
\begin{split}x_i^{(\lambda)} &=
\begin{cases}
\dfrac{x_i^\lambda - 1}{\lambda} & \text{if } \lambda \neq 0, \\[8pt]
\ln{(x_i)} & \text{if } \lambda = 0,
\end{cases}\end{split}
\label{eq:boxcox}
\end{equation}
where $\{\lambda\in\mathbb{R}\}$ is known as the power parameter determined through Maximum Likelihood estimation and $\mathbb\{x\in\mathbb{R}|x>0\}$
\begin{equation}
\begin{split}x_i^{(\lambda)} &=
\begin{cases}
 [(x_i + 1)^\lambda - 1] / \lambda & \text{if } \lambda \neq 0, x_i \geq 0, \\[8pt]
\ln{(x_i + 1)} & \text{if } \lambda = 0, x_i \geq 0 \\[8pt]
-[(-x_i + 1)^{2 - \lambda} - 1] / (2 - \lambda) & \text{if } \lambda \neq 2, x_i < 0, \\[8pt]
 - \ln (- x_i + 1) & \text{if } \lambda = 2, x_i < 0
\end{cases}\end{split}
\label{eq:yeo}
\end{equation}
Unlike the Box–Cox transform where $x_{i}$ must be positive, the Yeo–Johnson transform, given in Equation \ref{eq:yeo} holds for $\mathbb\{x\in\mathbb{R}\}$.

Before classification, the data is transformed feature-wise, i.e. the four features (percentages of speech, music, noise and silence) are individually "Gaussianized" by means of Z-score normalization and the Yeo–Johnson transform.

\subsection{Feature Learning \& Model Training}
\label{subsec:feat_learn}
\begin{figure}[t]
  \centering
  \includegraphics[width=0.2\textwidth]{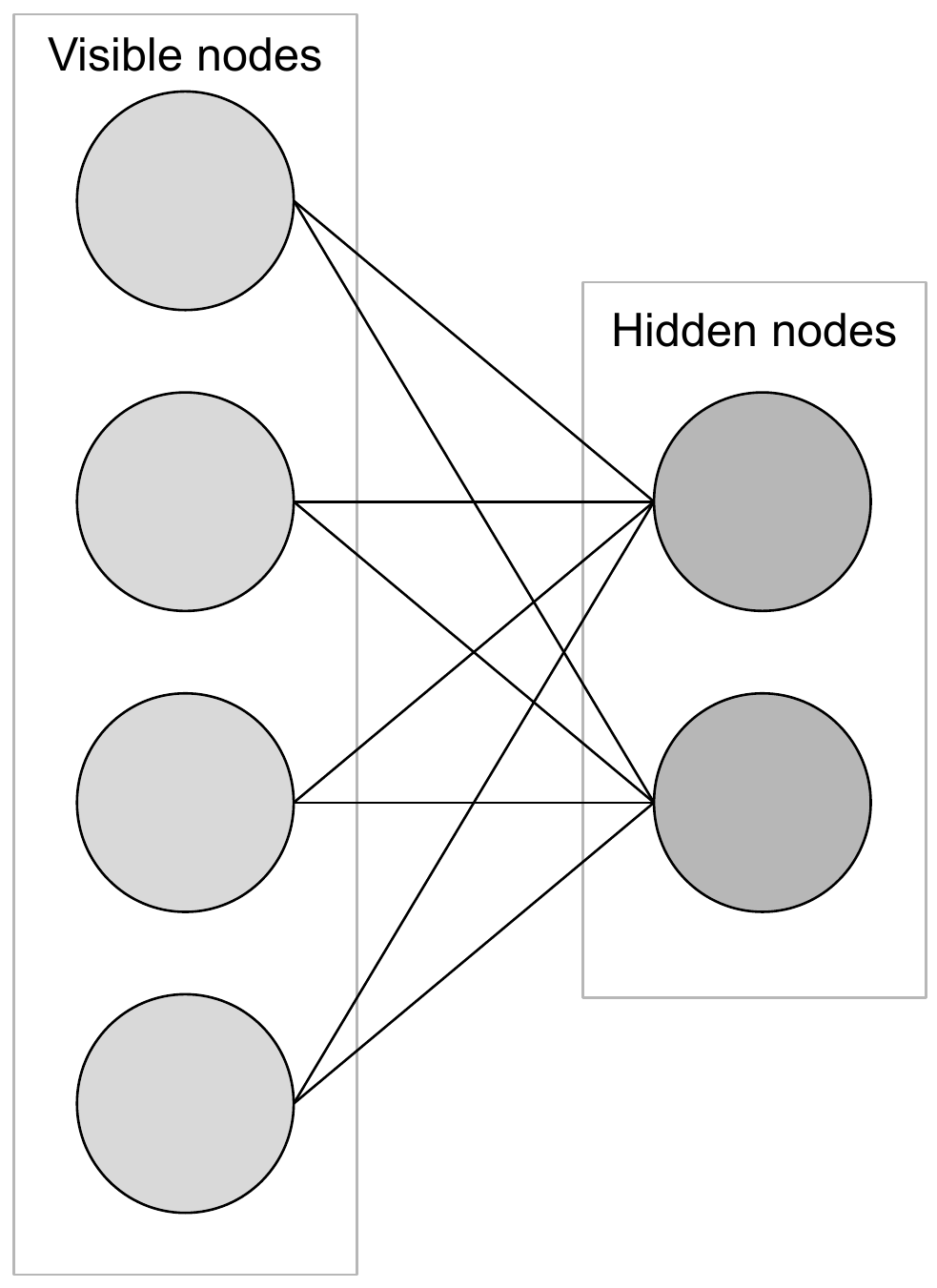}
  \caption{A restricted Boltzmann's machine (RBM) network with 4 visible nodes and two hidden nodes}
  \label{fig:rbm}
\end{figure}

To learn the best possible discriminative features, we employ a restricted Boltzmann's machine (RBM) network because of its relative simplicity and reported improved performance in the feature learning domain \citep{Salakhutdinov:2007}. Briefly, a RBM is a neural network that consists of a visible layer and one hidden layer with the restriction that the nodes must form a bipartite graph as depicted in Figure \ref{fig:rbm}. 

Training RBMs is conventionally done via gradient-based contrastive divergence algorithms \citep{Hinton:2002} which require practical knowledge about how to set the values of numerical meta-parameters such as the learning rate, the momentum, the weight-cost, the sparsity target, the initial values of the weights, the number of hidden units and the size of each mini-batch. In this study, the learning rate, number of hidden units and the size of the mini-batch were optimized via a grid search. The range of values selected for the grid search was guided by \citep{Hinton:2012} and are presented in Table \ref{tab:rbm_params}
\begin{table}[htbp]
\centering
\caption{RBM hyper-parameter grid values}
\label{tab:rbm_params}
\begin{tabular}{@{}ll@{}}
\toprule
Parameter           & Values                             \\ \midrule
No. of hidden nodes & {[}2, 20, 45, 70, 135, 170, 200{]} \\
Learning rate       & logspace(-3, 0, 20)                \\
Batch size          & {[}8, 16, 64, 128{]}               \\ \bottomrule
\end{tabular}
\end{table}

The RBM is cascaded with a k-means classifier to cluster the output into 2 groups, one for malpractices and the other for non-malpractices. One epoch of training entails passing the data through the RBM and clustering it then updating the parameters of the RBM based on the recall score obtained after clustering. Training stops after all the points on the hyperparameter grid have been tested and the set of hyperparameters that yield the highest score are selected.

\subsection{Experimental Set-up}
\label{subsec:exp_setup}
Our experiments were set up to systematically improve the performance of a preceding configuration. Three main experiments were conducted and the result of each was assessed with the recall score. In addition to the recall score, we compute the confusion matrix, the F1 score (Equations \ref{eq:precision} to \ref{eq:f1}) as well as the malpractice miss-classification error for each experiment.
\begin{align}
    \text{Precision} &= \frac{TP}{TP+FP}\label{eq:precision}\\
    \text{Recall} &= \frac{TP}{TP+FN}\label{eq:recall}\\
    \text{F1} &= 2\frac{Precision*Recall}{Precision+Recall}\label{eq:f1}
\end{align}
where $TP,\;FP,\;\&\;FN$ are true positive, false positive and false negative respectively. A value closer to 1 or 100\% is desirable in each case.

\begin{figure}[h]
  \centering
  \includegraphics[width=\textwidth]{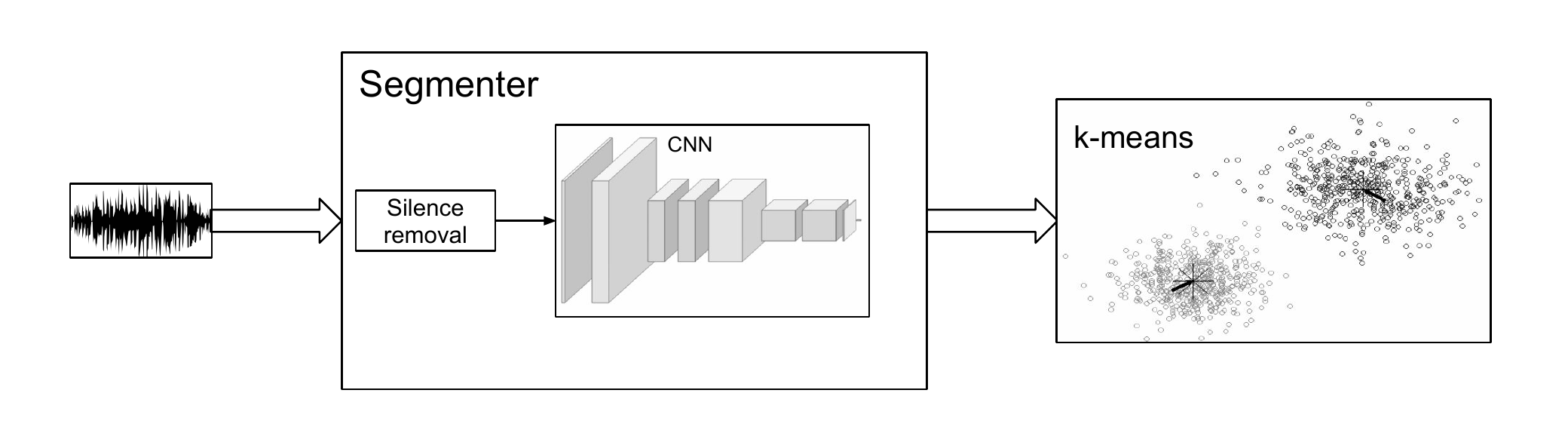}
  \caption{The extracted features fed into a k-means clustering model}
  \label{fig:kmeans}
\end{figure}

\begin{figure}[h]
  \centering
  \includegraphics[width=\textwidth]{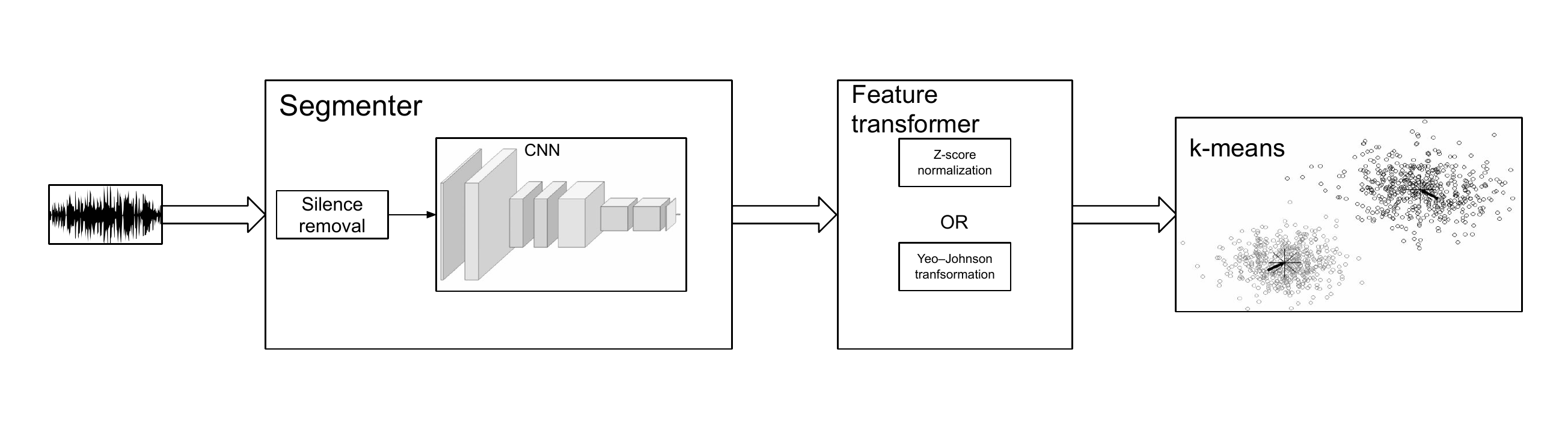}
  \caption{The extracted features transformed before clustering using the k-means model}
  \label{fig:trans_kmeans}
\end{figure}

\begin{figure}[h]
  \centering
  \includegraphics[width=\textwidth]{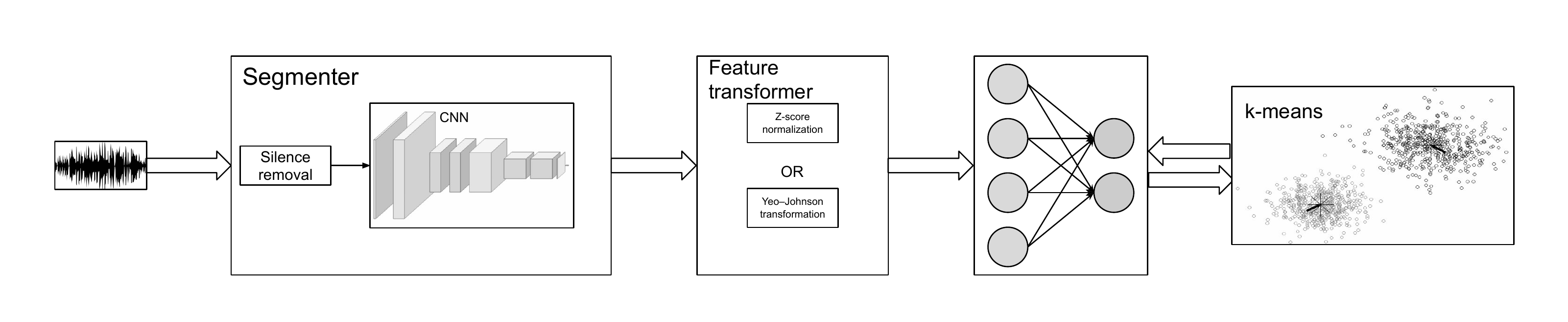}
  \caption{Experimental set-up for feature extraction and feature learning for k-means clustering}
  \label{fig:trans_rbm_kmeans}
\end{figure}

The three consecutive experimental set-ups can be summarized as follows:

\begin{itemize}
\item \textbf{k-means:}
For this experiment, we simply feed the extracted features as input to a k-means clustering models with two clusters representing malpractice and non-malpractice as shown in Figure \ref{fig:kmeans}.

\item \textbf{Feature transformation and k-means:}
In this experimental set-up, we built on top of the previous set-up by including the feature transformation block. Two sub-experiments are performed, one for Z-score normalized features and another for the Yeo–Johnson transformed features. The block diagram of Figure \ref{fig:trans_kmeans} depicts these experiments.

\item \textbf{Feature transformation, neural feature learning and k-means:}
Our final experiments included the feature learning block explained in Sub-section \ref{subsec:feat_learn}. We performed two sub-experiments for the two feature transformations and the parameters of the RBM were optimized in both cases. The procedure is summarized in Figure \ref{fig:trans_rbm_kmeans}.
\end{itemize}

\section{Results}\label{sec:results}
We present the results in three parts corresponding to the three main experiments performed in this study.
In each case, the model was trained with the unlabelled data and validated with the labeled data to assess the performance and tune the hyper-parameters as the case may be.

\subsection{K-means clustering}
\begin{figure}[h]
  \centering
  \includegraphics[width=0.5\textwidth]{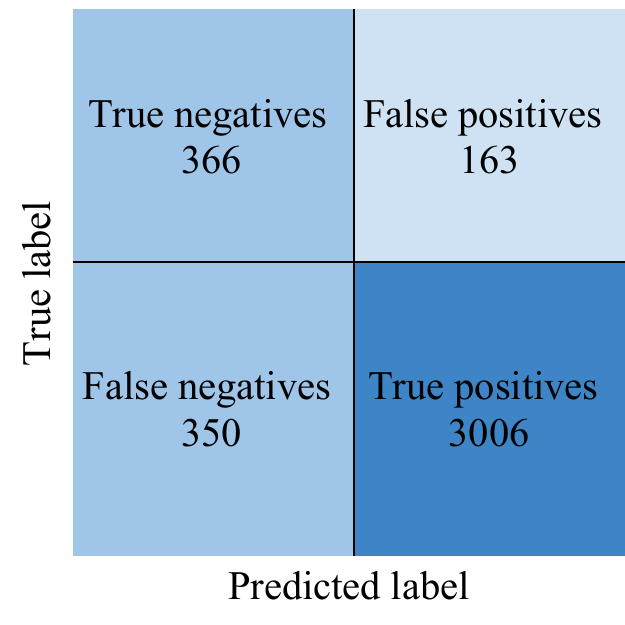}
  \caption{The confusion matrix obtained after k-means clustering}
  \label{fig:confmat_kmeans}
\end{figure}

K-means clustering was performed for two clusters on 176,115 samples. The rest of the samples (3885) were used as validation data to assess the performance of the model. Based on domain knowledge as to the distribution of the data, the cluster with more samples was assumed to be the cluster of non-malpractice calls while the smaller cluster corresponded to calls which constituted a form of malpractice. The confusion matrix obtained after k-means clustering are shown in Figure \ref{fig:confmat_kmeans}.

\subsection{Neural Feature Learning}
\begin{figure}[h]
  \centering
  \includegraphics[width=0.5\textwidth]{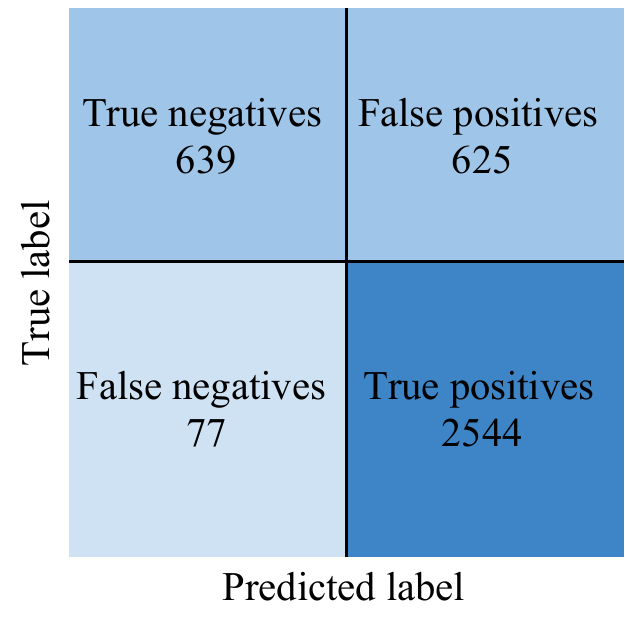}
  \caption{The confusion matrix obtained after Z-score normalization and incorporation of a RBM network for feature learning}
  \label{fig:confmat_rbm_ztrans_kmeans}
\end{figure}

\begin{figure}[h]
  \centering
  \includegraphics[width=0.5\textwidth]{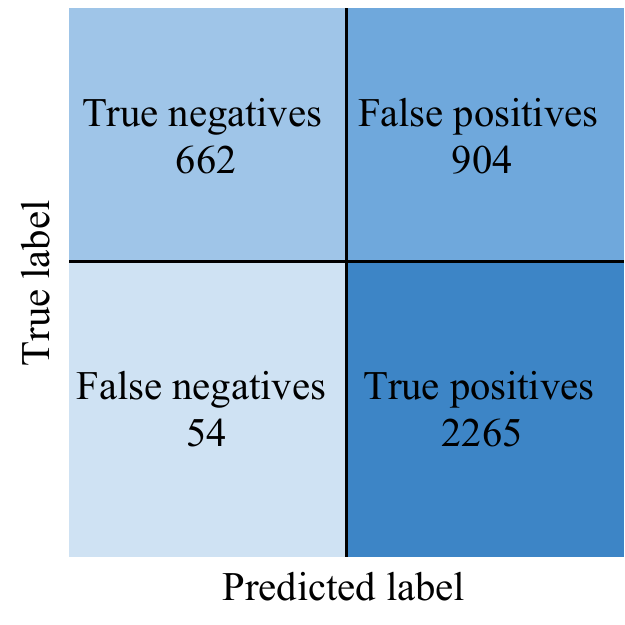}
  \caption{The confusion matrix obtained after Yeo–Johnson power transformation and incorporation of a RBM network for feature learning}
  \label{fig:confmat_rbm_ptrans_kmeans}
\end{figure}

Neural feature learning involves incorporating an RBM architecture to extract more discriminative features as seen in Figure \ref{fig:trans_rbm_kmeans}. The parameters of a RBM network are tuned to maximize the recall score. In Table \ref{tab:recall_f1}, the optimal parameters of the network as well as the performance scores are presented. Similar to the experiment carried out in Sub-section \ref{ssec:trans_kmeans}, the performance scores were computed for Z-score normalized data and Yeo–Johnson power transformed data. The confusion matrices are presented in Figures \ref{fig:confmat_rbm_ztrans_kmeans} and \ref{fig:confmat_rbm_ptrans_kmeans}.

\subsection{Feature Transformation \& K-means clustering}
\label{ssec:trans_kmeans}
\begin{figure}[h]
  \centering
  \includegraphics[width=0.5\textwidth]{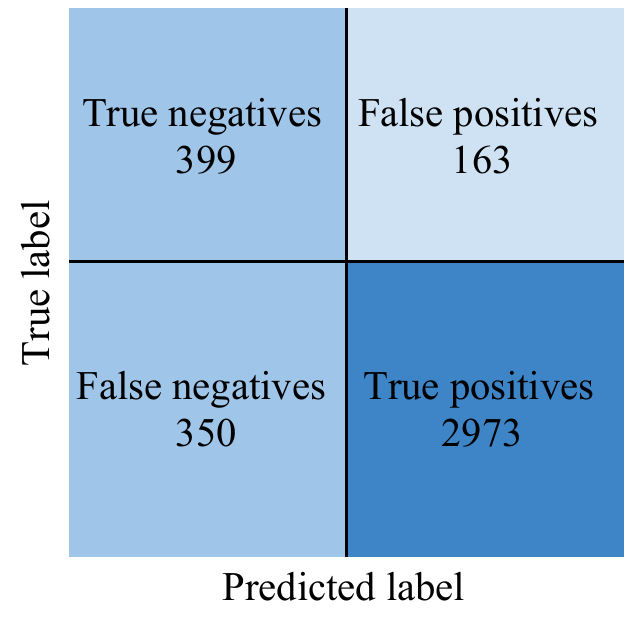}
  \caption{The confusion matrix obtained after Z-score normalization and k-means clustering}
  \label{fig:confmat_ztrans_kmeans}
\end{figure}

\begin{figure}[h]
  \centering
  \includegraphics[width=0.5\textwidth]{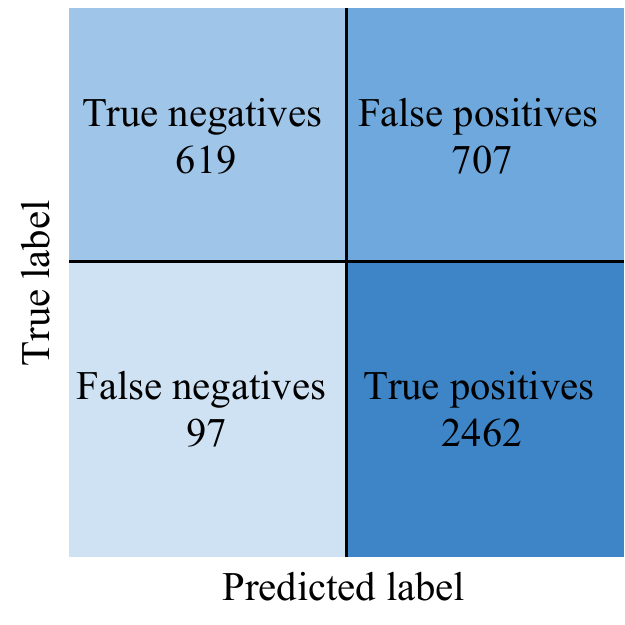}
  \caption{The confusion matrix obtained after Yeo–Johnson power transformation and k-means clustering}
  \label{fig:confmat_ptrans_kmeans}
\end{figure}

In Figures \ref{fig:confmat_ztrans_kmeans} and \ref{fig:confmat_ptrans_kmeans}, we present the confusion matrices obtained from the feature transformations and k-means clustering experiments. Specifically, Z-score normalization and Yeo–Johnson power transformation were performed before clustering with the k-means algorithm.

The results presented in this section are obtained from the experimental set-up of Figure \ref{fig:kmeans}.

Finally, the recall and F1 scores as well as the optimal parameters of the RBM networks are presented in Table \ref{tab:recall_f1}.

\begin{table}[]
\centering
\caption{Recall and F1 scores obtained for the five models as well as the optimal parameters for training the RBM. ZN=Z-score normalization, PT=power transformation, RBM=restricted Boltzmann's machines}
\label{tab:recall_f1}
\resizebox{\textwidth}{!}{%
\begin{tabular}{cccccl}
\hline
\multicolumn{1}{l}{Recall} & \multicolumn{1}{l}{F1} & \multicolumn{1}{l}{Batch size} & \multicolumn{1}{l}{Learning rate} & \multicolumn{1}{l}{Hidden units} & Model \\ \hline
0.871          & 0.903 & -  & -     & -  & k-means                   \\
0.948          & 0.921 & -  & -     & -  & ZN\_k-means               \\
0.962          & 0.86  & -  & -     & -  & PT\_k-means               \\
0.971          & 0.879 & 64 & 0.113 & 20 & ZN\_RBM\_k-means          \\
\textbf{0.977} & 0.825 & 8  & 0.004 & 2  & \textbf{PT\_RBM\_k-means} \\ \hline
\end{tabular}%
}
\end{table}

\section{Discussion}
In this section, we interpret the results, provide possible explanations for them and outline the implications in a real world scenario. All the experiments yielded good results, above 80\% for each measure, yet there was room for improvements. For this use case, the aim is to minimize the false negatives and maximize the recall score.

For the k-means-only model, a recall score of 0.896 was obtained with an F1 score of 0.92. While these scores are reflective of a good model, a closer examination of the confusion matrix reveals that  almost 50\% of the actual malpractices were classified as non-malpractice. This means that these calls will go unchecked by the quality control personnel. Thus, it was imperative to improve on this result by trying to minimize the number of wrongly classified malpractice calls. We call this the Malpractice Classification Error (MCE).

By ``Gaussianizing''  the data before clustering, we obtained a more desirable performance. Z-score normalization did not result in any significant change however, power transformation yielded an improved recall score of 0.962. The F1 scores on the other hand, decreased. This is as a result of the trade-off between precision and recall. We also observed a significant improvement in performance in terms of the MCE for the power transformed data. Specifically, the MCE reduced from 48\% to about 13.5\%.

To further improve on the results obtained after the addition of data transformations, we applied the so called neural feature learning. A significant increase in the recall scores can be observed in Table \ref{tab:recall_f1}. While the recall scores for both Z-score normalization and power transformation combined with neural feature learning are higher than the previously discussed results; that of the power transformation is more significant. Concretely, after hyper-parameter tuning, we obtained a recall score of 0.977. The MCE was reduced to about 7.5\%, showing significant improvement from the previous results.

Regarding the parameters of the RBM, the optimum values, in this case affect the speed of training. Where a smaller batch size and learning rate translate to slower training of the RBM when compared to larger values. Conversely, a large number of hidden units translates to slower overall training since the k-means classifier will have to work on a higher dimension of clustering data. A clear trade-off can be observed between performance and training speed.

\begin{figure}[]
  \centering
  \includegraphics[width=\textwidth]{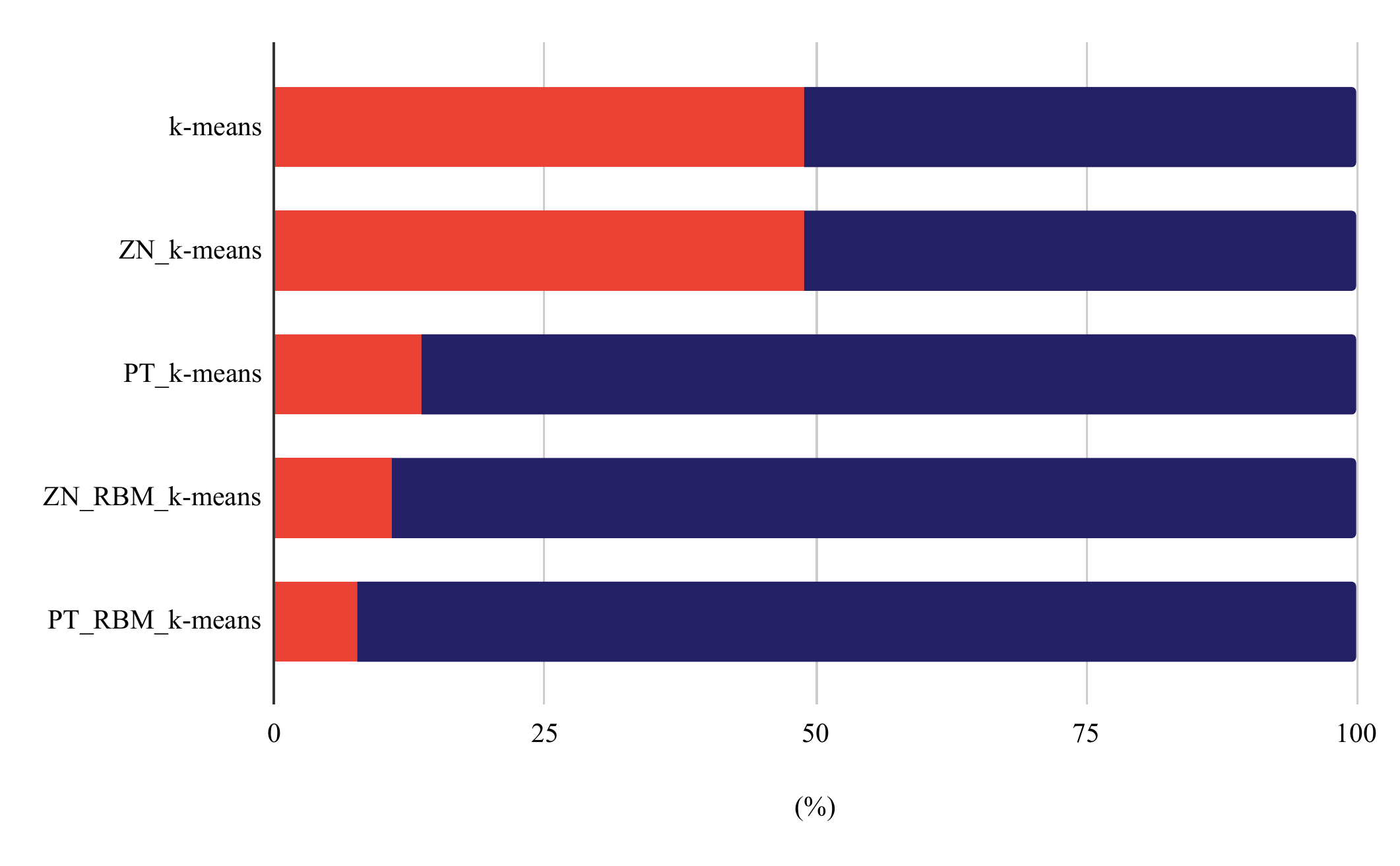}
  \caption{Malpractice classification error (in red) of the various models trained in this study}
  \label{fig:mce}
\end{figure}

The decreasing trend of the MCE seen in Figure \ref{fig:mce} is desirable as it shows progressive improvement in the models that we experimented with. Although a higher F1  score is coveted, it will be difficult to achieve since we maximized the recall score. Nevertheless, an F1 score of over 0.8 is considered good.

\begin{figure}[]
  \centering
  \includegraphics[width=\textwidth]{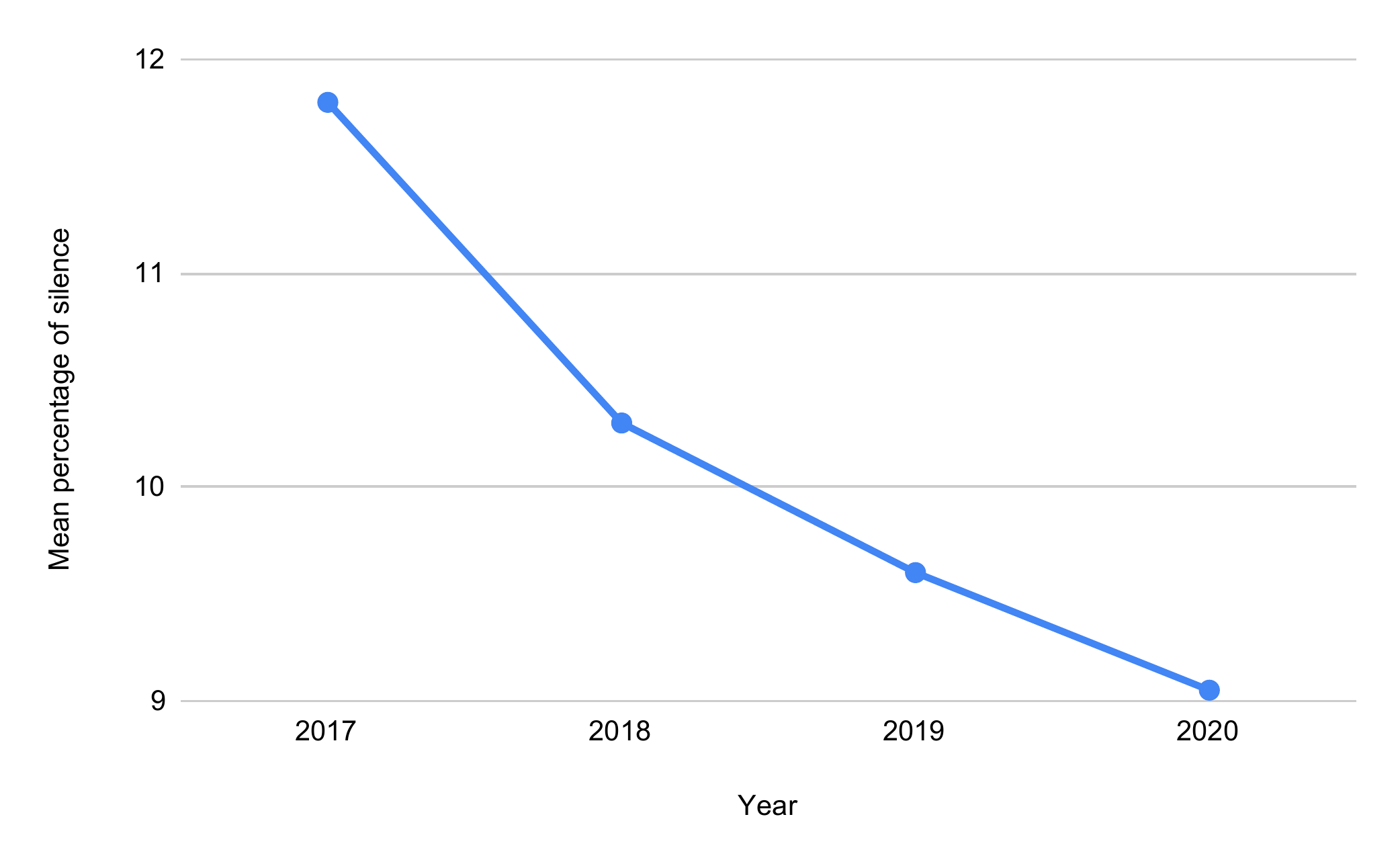}
  \caption{The mean percentage of silence of calls per year at our in-house call center}
  \label{fig:avg_silence}
\end{figure}

Finally, at the onset of incorporating the automatic malpractice detection system into our call center, we observed a sharp decrease in the duration of silence in calls (Figure \ref{fig:avg_silence}). As the quality control personnel become more familiar with the system, they have been improving on their feedback and this has seen further decrease in silences in calls.

\section{Conclusions and Future Work}\label{sec:conc}
In this work, we have shown that by careful feature engineering, it is possible to improve the performance of a simple k-means algorithm. The system leveraged the hyperparameter tuning of a RBM to obtain the best possible model performance. We applied the method, in a semi-supervised manner, to call center agent malpractice detection where we sought to maximize the recall score and minimize the MCE.

We found that the non-linear, power transformation outperformed the more widely used Z-score normalization in the considered scenarios. With an F1 score of 0.825 and a recall score of 0.977, the model is sufficiently good enough to be deployed in a real world scenario as is the case in our in-house call center. The corresponding MCE of  7.54\% will greatly impact the efficiency of the quality control personnel since their efforts towards detecting agent malpractices can be more focused with a small margin of error.

Since the mean amount of silence detected in an agent's calls over a period of time may be used as one of the key performance indicators, at our call center, we monitor it and consequently, we have reported a decrease in this value since the implementation of the automatic agent malpractice detection system.

Our study is not without limitations. Firstly, the size of the validation data may be increased so as to have a more robust system. Additionally, rather than performing a grid search for the optimal hyperparameters of the RBM, a random search would produce more desirable results. We did not explore optimizing the number of iterations at training though a significant improvement in performance is not expected. We use a constant learning rate for training however, in \citep{Tieleman:2008}, it was noted that in practice, decaying learning rates often work better.

The future work of this study points in various directions but of particular interest is distinguishing the types of agent malpractices. This may intuitively be achieved by defining more clusters for k-means or by employing some of the methods used to obtain the optimum number of clusters in a data set. Since the system is made up of modules, it allows for the flexibility of substituting the various modules while optimizing the performance. Concretely, in the literature, there are similar transformation algorithms that may be used instead of the Yeo–Johnson power transform. The neural network may be replaced with a more sophisticated architecture such as autoencoders or networks comprising of a combination of one-dimensional convolutional layers and fully connected layers.

\noindent
Funding: This work was supported by the Scientific and Technological Research Council of Turkey (TÜBİTAK), Grant No: TEYDEB1507-7170694.

\bibliography{mybibfile}
\end{document}